\documentclass{article}
\usepackage{ijcai25}

\usepackage{times}
\usepackage{soul}
\usepackage{url}
\usepackage[hidelinks]{hyperref}
\usepackage[utf8]{inputenc}
\usepackage[small]{caption}
\usepackage{graphicx}
\usepackage{amsmath}
\usepackage{amsthm}
\usepackage{booktabs}
\usepackage{algorithm}
\usepackage{algorithmic}
\usepackage[switch]{lineno}

\usepackage{color}
\usepackage{amssymb}
\usepackage{moresize}
\usepackage{subfigure}
\usepackage{xcolor}

\renewcommand{\textcolor}[2]{#2}  
\usepackage{array}

\title{GATES: Cost-aware Dynamic Workflow Scheduling via Graph Attention Networks and Evolution Strategy}

\author{
Ya Shen\footnote{Corresponding author}
\and
Gang Chen\and
Hui Ma\And
Mengjie Zhang\\
\affiliations
Centre for Data Science and Artificial Intelligence \& School of Engineering and Computer Science,\\
Victoria University of Wellington, New Zealand \\
\emails\{ya.shen, aaron.chen, hui.ma, mengjie.zhang\}@ecs.vuw.ac.nz
}

\begin{document}

\maketitle

\begin{abstract}
Cost-aware Dynamic Workflow Scheduling (CADWS) is a key challenge in cloud computing, focusing on devising an effective scheduling policy to efficiently schedule dynamically arriving workflow tasks, represented as Directed Acyclic Graphs (DAG), to suitable virtual machines (VMs). Deep reinforcement learning (DRL) has been widely employed for automated scheduling policy design. However, the performance of DRL is heavily influenced by the design of the problem-tailored policy network and is highly sensitive to hyperparameters and the design of reward feedback. Considering the above-mentioned issues, this study proposes a novel DRL method combining \textbf{G}raph \textbf{A}ttention Ne\textbf{t}works-based policy network and \textbf{E}volution \textbf{S}trategy, referred to as \textbf{GATES}. The contributions of GATES are summarized as follows: (1) GATES can capture the impact of current task scheduling on subsequent tasks by learning the topological relationships between tasks in a DAG. (2) GATES can assess the importance of each VM to the ready task, enabling it to adapt to dynamically changing VM resources. (3) Utilizing Evolution Strategy's robustness, exploratory nature, and tolerance for delayed rewards, GATES achieves stable policy learning in CADWS. Extensive experimental results demonstrate the superiority of the proposed GATES in CADWS, outperforming several state-of-the-art algorithms. The source code is available at: https://github.com/YaShen998/GATES.
\end{abstract}

\section{Introduction}\label{sec:introduction}
Real-world problems are often highly complex, requiring applications tailored to solve them to execute efficiently and deliver prompt results to meet user demands~\cite{pallathadka2022investigation}. For example, if a weather forecasting application fails to run in real time and provide timely, accurate predictions, it may interrupt the regular management of vital industrial operations~\cite{price2025probabilistic}. Cloud computing enables the deployment of such complex applications structured as workflows consisting of multiple tasks. It facilitates the scheduling of elastic computational resources to ensure efficient and timely task execution~\cite{jadeja2012cloud}. This process is known as Workflow Scheduling (WS).

A workflow can be represented as a \emph{Directed Acyclic Graph} (DAG), with vertices representing tasks and edges representing task dependencies~\cite{shen2024cost}. Workflows with different patterns have different DAG structures. Additionally, each workflow includes a \emph{Service Level Agreement} (SLA) that defines the maximum allowed completion time for the workflow~\cite{qazi2024service}. This study focuses on the Cost-aware Dynamic Workflow Scheduling (CADWS) problem~\cite{shen2024cost}, which aims to develop an efficient scheduling policy that minimizes VM rental costs and SLA violation penalties, while handling the complexity caused by the dynamic arrival of workflows with diverse DAG structures~\cite{huang2022cost}.

Compared to traditional scheduling problems such as job shop scheduling and vehicle routing, CADWS  exhibits complex dynamic characteristics, including unpredictable workflow arrivals, varying workflow patterns, and rapidly fluctuating computational resources (e.g., the number of available VMs). Moreover, unlike job shop and vehicle routing problems, workflows in CADWS have highly diverse graph topologies across different patterns, while the others maintain similar structures regardless of scale. This study focuses on using deep reinforcement learning (DRL) techniques to automatically learn scheduling policies for solving CADWS, with an emphasis on designing effective policy networks tailored to the problem characteristics of CADWS.

The key challenges of this study are: (1) how to design an effective policy network to properly control the influence of every scheduling decision on future task execution in a dynamic environment with continuously arriving workflows of highly diverse patterns; (2) how can the designed policy network support a time-varying collection of VMs; and (3) how to reliably train this policy network to ensure robust performance across varying system settings. To address the aforementioned challenges, this paper proposes a novel DRL method that integrates \textbf{G}raph \textbf{A}ttention Ne\textbf{t}works (GAT) and \textbf{E}volution \textbf{S}trategy (ES), referred to as \textbf{GATES}.

Specifically, the main contributions of GATES are as follows: (1) GATES employs GAT to efficiently and scalably process the local topological information of each ready task. Leveraging GAT's attention mechanism, GATES accurately learns the global attention information between the ready task and other tasks, enabling a holistic view of current scheduling impacts on future tasks by aggregating all task embeddings. (2) Dynamic heterogeneous graphs with analogous structures but varying scales are constructed to represent the relationships between the ready task and all available VMs at each time step. Regardless of dynamic changes in VM numbers, GAT can compute attention coefficients from these graphs to identify the most suitable VM at each time step. (3) In CADWS, the reward function typically includes VM rental fees and SLA violation penalties. However, since SLA penalties can only be accurately determined after workflow execution, it’s difficult to design a reward function that quantifies SLA violations at each time step. To overcome this challenge, ES is utilized to train the policy network. This approach eliminates the need for complex reward design and hyperparameter tuning, while also preventing performance issues caused by poorly designed rewards.
\section{Related Work}\label{sec:related_works}
Currently, most methods for solving scheduling problems focus on designing effective scheduling policies~\cite{branke2015automated,khalil2017learning,dong2021predictive,hasanzadezonuzy2021model}. Some manually designed policies have been developed to solve CADWS~\cite{pham2020evolutionary,wu2017deadline,faragardi2019grp}, as these methods are easy to understand and implement. However, these methods heavily rely on expert experience and substantial domain knowledge, limiting their adaptability to dynamic environments~\cite{geiger2006rapid}, as they are developed based on historical workflows and may fail to perform effectively in rapidly changing scenarios or with unknown workflows~\cite{shen2024cost}. 

Several DRL studies~\cite{kayhan2023reinforcement,cunha2020deep,ou2023deep,iklassov2023study} have shown that using neural network-based policy networks to automatically learn scheduling policies is a feasible and promising approach to solving scheduling problems. For instance,  \citeauthor{ou2023deep}~[\citeyear{ou2023deep}] utilized DRL to solve the satellite range scheduling problem, while Song et al.~[\citeyear{song2022flexible}] and Zhang et al.~[\citeyear{zhang2020learning}] employed DRL to learn effective policies for addressing the job-shop scheduling problem. However, the problems addressed by these works are either static or do not have a graph topological structure, such as the DAG of CADWS. Moreover, although \citeauthor{shen2024cost}~[\citeyear{shen2024cost}] and  \citeauthor{huang2022cost}~[\citeyear{huang2022cost}] applied DRL to dynamic WS, their policy networks ignored the DAG structures of workflows, failing to leverage task dependencies for effective scheduling. Thus, the architecture of policy networks designed in these approaches did not take dynamic information or graph topological information into account to extract essential knowledge for learning scheduling policies, making them unsuitable for CADWS. 
   
Recently, some studies~\cite{sun2023difusco,zhang2024deep} proposed graph-based policy networks to extract state features for effective policy learning by leveraging the graph topological properties inherent in problems, such as traveling salesman, maximal independent set, and job shop problems. \citeauthor{sun2023difusco}~[\citeyear{sun2023difusco}] proposed a graph-based diffusion model to exploit the graph topological information in the traveling salesman problem and the maximal independent set problem. Similarly, \citeauthor{zhang2024deep}~[\citeyear{zhang2024deep}] utilized Graph Neural Networks (GNNs) to extract state features for effective policy learning by leveraging the graph topological properties of the job shop scheduling problems. While the above studies considered the graph topological information, they assume static structure and fixed scheduling resources. In contrast, CADWS accounts not only for the dynamic workflow arrivals and patterns but also for the variability in scheduling resources over time~\cite{shen2024cost}. Specifically, the pool of active VMs can change across different system states and time steps.

Unlike existing DRL approaches, GATES is specifically designed in this study to address the dynamic nature of scheduling problems. It handles the unpredictable arrival and varying patterns of workflows, as well as fluctuations in available scheduling resources, achieving stable policy learning in complex and dynamic environments.

\section{Problem Definition}\label{sec:problem_definition}
\subsection{Preliminary of CADWS}

The CADWS problem involves a set of workflows $ \mathcal{W} $ arriving dynamically over time and a dynamically changing collection of heterogeneous VMs $ \mathcal{M} $ for task execution, meaning that $|\mathcal{M}| $ is not fixed. Each workflow $ W_i \in \mathcal{W} $ is modeled as a DAG $ ( \mathcal{O}_{W_i}, \mathcal{C}_{W_i} ) $, where the node set $ \mathcal{O}_{W_i} = \{ O_{i1}, O_{i2}, \dots, O_{in} \} $ represents tasks, and the edge set $ \mathcal{C}_{W_i} = \{ (O_{ij}, O_{ik}) \mid O_{ij}, O_{ik} \in \mathcal{O}_{W_i} \} $ denotes precedence constraints (see Appendix A). An edge $ (O_{ij}, O_{ik}) $ implies that task $ O_{ij} $ must complete before task $ O_{ik} $ can begin execution. $ O_{ij} $ becomes a ready task $ O^{*} $ waiting for current scheduling when there are no predecessor tasks for it~\cite{zhu2016fault}.

The complete set of tasks across all workflows is denoted as $ \mathcal{O} = \mathcal{O}_{W_1} \cup \mathcal{O}_{W_2} \cup \dots \cup \mathcal{O}_{W_{|\mathcal{W}|}} $. Each task $ O_{ij} $ is associated with a workload $ WL_{ij} \in \mathbb{R}^+ $ and can be executed on any VM $ M_k \in \mathcal{M} $. The execution time of $ O_{ij} $ on VM $ M_k $ is determined as $ ET(O_{ij})^{k} = \frac{WL_{ij}}{PS_{k}} $, where $ PS_{k} $ denotes the processing speed of machine $ M_k $. Hence, $ CT(O_{ij}) = ST(O_{ij}) + ET(O_{ij})^{k}$ gives the completion time of $ O_{ij} $ on $ M_k $, where $ ST(O_{ij}) $ is the start time of $ O_{ij} $. $ O_{ij} $ is the start task of $ W_i $, if $ ST(O_{ij}) = AT(W_i) $, where $ AT(W_i) $ is the arrival time of workflow $ W_i $. The finish time of workflow $ W_i $ is defined as $ FT(W_i) = \max_j\{CT(O_{ij})\} $. The total process time of workflow $ W_i $ is computed as $ PT(W_i) = FT(W_i) - AT(W_i) $.

\subsection{Optimization Objective of CADWS}
The VM rental fees cover the total cost of all leased VMs used to execute all workflows dynamically arrived during a given time period $T$. This period, starting at time $t_s$ and ending at $t_e$, depends on the scheduling policy $\pi$ adopted. For each VM $v \in \mathcal{M}$, its usage period during $T$ is denoted as: $ UP(v, \pi, T) = \left[t_s(v, \pi, T), t_e(v, \pi, T)\right] $, where $t_s(v, \pi, T)$ represents the start time of using $v$, and $t_e(v, \pi, T)$ indicates the time when $v$ is no longer in use, satisfying $t_e(v, \pi, T) \leq t_e$. The total rental cost for all VMs under the scheduling policy $\pi$ over $T$ is computed as: $ VM_{Fee}(\pi, T) = \sum_{v \in Set(\pi, T)} \left(Price(v) \times \left\lceil \frac{t_e(v, \pi, T) - t_s(v, \pi, T)}{3600} \right\rceil \right) $, where $Set(\pi, T)$ is the set of VM instances leased for task execution during $T$. The ceiling function ensures that the rental time during an hour will be charged as an hour, following the common practice in the cloud computing industry~\cite{shen2024cost,huang2022cost}.

The SLA penalty for a workflow $w_i$ under a scheduling policy $\pi$ is defined as: $ SLA_{Penalty}(W_i, \pi) = \beta \times \max\{0, [WCT(W_i, \pi) - DL(W_i)]\} $, where $WCT(W_i, \pi)$ refers to the $PT(W_i)$ by following policy $\pi$, and $DL(W_i)$ denotes the SLA-specified deadline for $W_i$. The penalty factor $\beta$ quantifies the severity of SLA violations. Following many existing studies~\cite{huang2022cost,shen2024cost}, the deadline $DL(W_i)$ is calculated as: $ DL(W_i) = AT(W_i) + \gamma \times MinMakespan(W_i) $, where $AT(W_i)$ is the workflow's arrival time, $MinMakespan(W_i)$ is the theoretical minimum time required to process $W_i$ using the fastest VMs without delays, and $\gamma$ is a relaxation coefficient. A higher $\gamma$ value allows for a more lenient deadline.

The objective of CADWS is to find an optimal scheduling policy $\pi$ that minimizes the total cost, including both the VM rental fees and SLA penalties, as formulated below:
\begin{align}
\label{Objective function}
\arg\min_{\pi} TotalCost(\pi) = & \arg\min_{\pi} \{VM_{Fee}(\pi, T)  \\
& + \sum_{W_i \in \mathcal{W}} SLA_{Penalty}(W_i, \pi) \} \nonumber 
\end{align}
\section{Methology}\label{sec:methology}
\begin{figure}[htbp]
\centering
\includegraphics[width=0.9\linewidth]{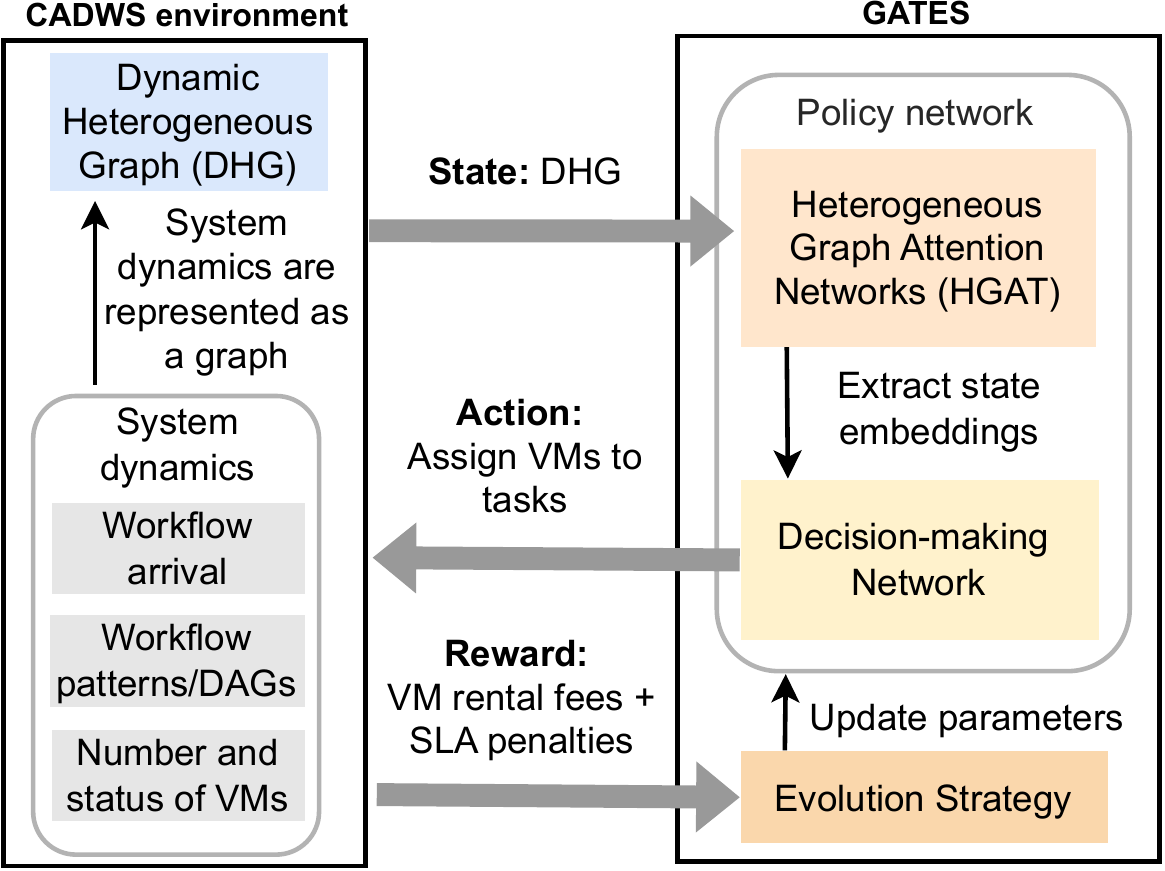} 
\caption{The overall framework of CADWS.}
\label{CADWS_framework}
\end{figure}

This section elaborates on the proposed GATES for solving CADWS in detail. CADWS is a sequential decision-making process that can be formulated as a Markov Decision Process (MDP)~\cite{huang2022cost,shen2024cost,yanggraph}, where scheduling decisions are made iteratively. At each step, GATES selects a compatible VM to execute a specific ready workflow task, continuing this process until all workflow tasks are scheduled. The workflow of GATES for solving CADWS is illustrated in Figure~\ref {CADWS_framework}.

As shown in Figure~\ref{CADWS_framework}, the current state of the scheduling system is captured by a Dynamic Heterogeneous Graph (DHG) representation (see Figure~\ref{dynamic_graph}). Subsequently, a heterogeneous Graph Attention Network (HGAT) employing a two-stage embedding mechanism is applied to process this graph. This step extracts feature embeddings of both tasks and VMs (see Figure~\ref{HGAT}). These embeddings are then utilized by a decision-making network to compute the action probability distribution, from which a VM is sampled or greedily selected (as detailed in Figure~\ref{decision_networks}). The next subsection introduces the MDP formulation of the CADWS  process.

\subsection{MDP Formulation of CADWS}
\textbf{System State}: Each system state $ s_t \in S $ represents a snapshot of the current state of the entire CADWS problem at decision step $ t $, including the following elements: (1) the current ready task $ O^* $ at $ t $; (2) details of all active workflows, encompassing pattern information on their constituent tasks and task dependencies (modeled as DAG), and workflow arrival times; (3) the numbers and processing status of all available VMs at $ t $. By designing an effective policy network that accurately captures the three aforementioned factors, the system state of CADWS at any time step can be precisely represented, addressing the first two challenges mentioned in Section~\ref{sec:introduction}. To achieve this, we define the system state of CADWS at any time step as a DHG (as shown in Figure~\ref{dynamic_graph}) and propose the HGAT employing GAT with a two-stage embedding mechanism to extract useful state embeddings for ready tasks (details in Figure~\ref{HGAT}), enabling the learning of effective scheduling policies and making informed decisions.

\textbf{Actions}: At the decision step $ t $, each action $ a_t \in \mathcal{A} $ assigns the ready task $ O^* $ to the waiting queue of an eligible VM $ M$ ($ M \in \mathcal{M} $). The size of the action space is $ |\mathcal{A}| = |\mathcal{M}| $.

\textbf{Transition}: Upon taking an action $ a_t $, the system transitions from state $ s_t $ to the next state $ s_{t+1} $, where the focused ready task at decision step $ t $ is assigned to a specific VM for execution (Refer to Figure~\ref{CADWS_framework}).

\textbf{Rewards}: The reward $ r_t $ provides a scalar feedback signal based on the action $ a_t $ taken at system state $ s_t $. To achieve the objective of minimizing the total cost as formulated in Eq.~\eqref{Objective function}, the reward is defined as $ r_t = - (VM_{Fee} +  SLA_{Penalty}) $.

\textbf{Policy}: The policy $ \pi(a_t \mid s_t) $ of GATES defines a probability distribution over the action set $ \mathcal{A} $ for each state $ s_t $, modeled as a parametric neural network. 

\textbf{Learning Objective}: The goal of the scheduling agent is to learn an optimal policy that maximizes the total rewards after the completion of the CADWS problem.

\subsection{Dynamic Heterogeneous Graph (DHG)}
In CADWS, we leverage the topological information of the workflows represented as DAGs to learn embeddings for ready tasks. These embeddings not only capture the features of the current ready task but also learn the global features by considering the relationships with other tasks. This DAG-based global feature learning reveals the influence of predecessor tasks on the ready task and the potential impact of the ready task on subsequent tasks.

\begin{figure}[htbp]
\centering
\includegraphics[width=\linewidth]{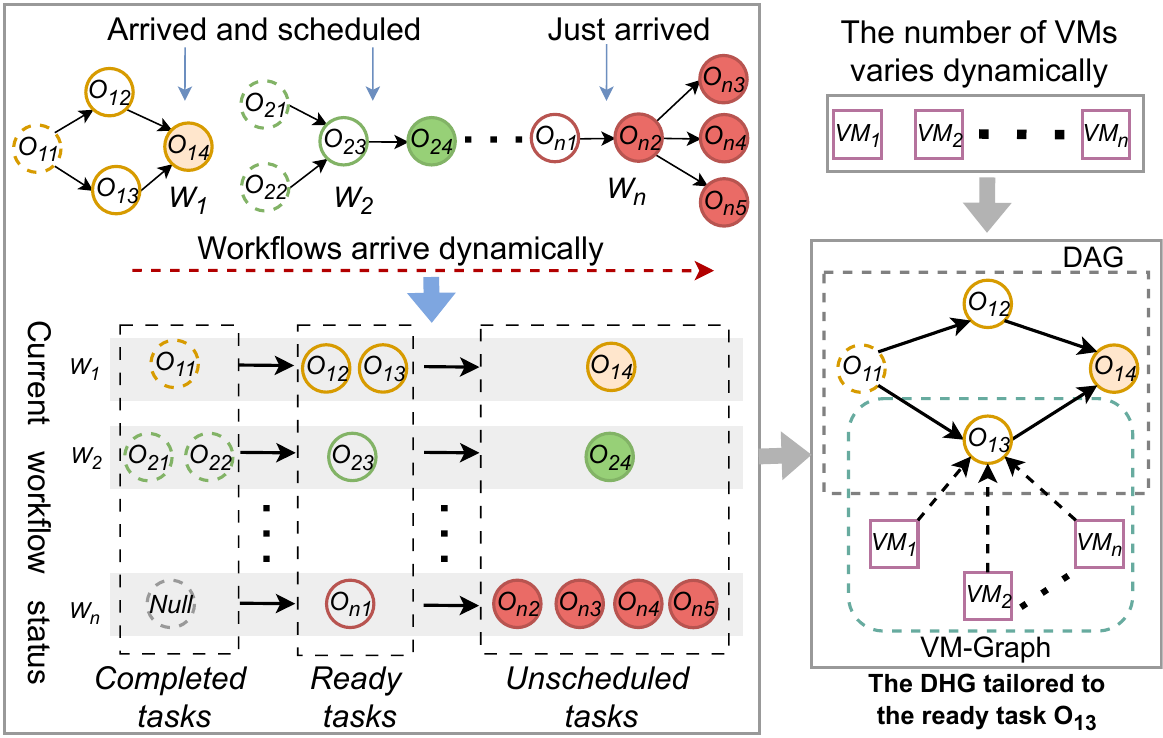} 
\caption{The proposed Dynamic Heterogeneous Graph (DHG).}
\label{dynamic_graph}
\end{figure}

Additionally, we incorporate dynamically changing VMs into the DAG to construct a DHG, as shown in Figure~\ref{dynamic_graph}. Since the number and types of VMs vary dynamically at different decision steps, the VM nodes and edges in the DHG are inherently dynamic. The DHG provides a global perspective to describe the influence of all tasks and currently available VMs on the scheduling of the ready task. Moreover, the DHG is constructed specifically for each ready task: whenever a task becomes ready and awaits scheduling, a DHG is generated to capture the essential system's dynamics for executing the ready task.

\subsection{Proposed Policy Network}
To support scheduling policy learning in GATES (the Agent), it is crucial to extract state embedding from the constructed DHG. GNNs are well-suited for learning embeddings from graph data. Among them, GAT outperforms Graph Convolutional Networks (GCN) and Graph Isomorphism Networks (GIN) in learning node embeddings from heterogeneous graphs. This advantage arises from GAT's use of attention mechanisms to assign dynamic weights to each neighboring node, enabling it to adapt to heterogeneous node features and diverse neighborhood relationships~\cite{velivckovic2017graph}. In contrast, GCN and GIN rely on fixed weights or mean aggregation, limiting their ability to model feature differences in heterogeneous graphs~\cite{thekumparampil2018attention}. Moreover, GAT does not require strict alignment of node features and naturally handles various relationships in heterogeneous graphs through its multi-head attention mechanism, offering greater flexibility and scalability~\cite{wang2019heterogeneous}. 

\subsubsection{Heterogeneous Graph Attention Networks (HGAT)}
\begin{figure}[htbp]
\centering
\includegraphics[width=\linewidth]{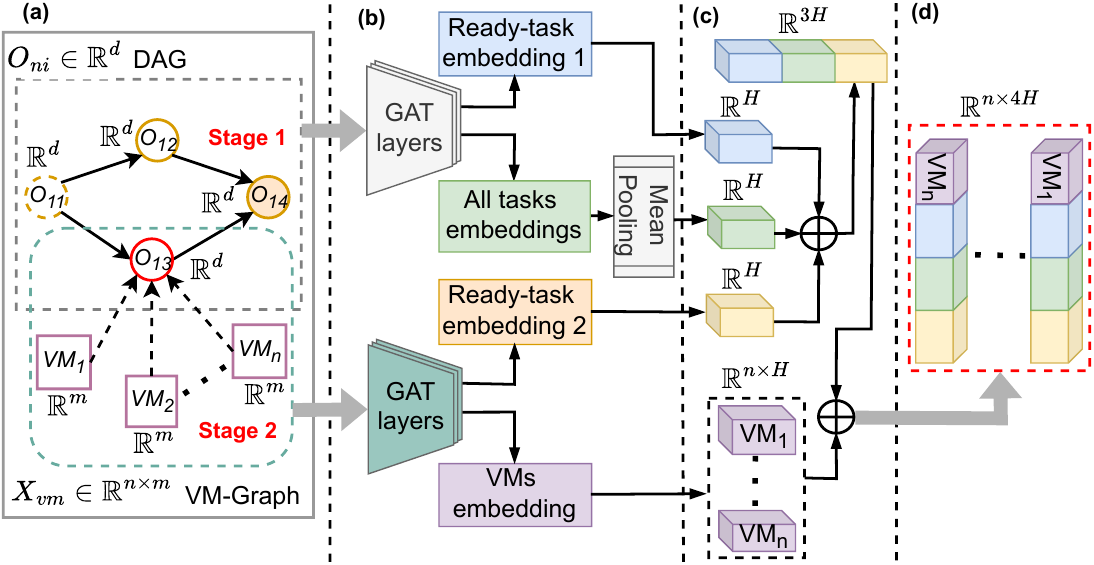} 
\caption{The proposed HGAT: (a) a DHG tailored to the ready task, (b) the embedding learning process, (c) feature concatenation, (d) output the learned state embedding.}
\label{HGAT}
\end{figure}

We propose an HGAT to extract the state embedding from the DHG, as shown in Figure~\ref{HGAT}. The embedding learning process of HGAT consists of two stages: (1) Use GAT to learn feature embedding for each task from the DAG (Stage 1 of Figure~\ref{HGAT}), in which the \textbf{Ready-task embedding 1} refers to the feature embedding of the ready task. Meanwhile, the feature embeddings of all tasks are aggregated using mean pooling to obtain the \textbf{All tasks embedding}; (2) Another GAT is employed to learn the \textbf{Ready-task embedding 2} from the VM-graph in the DHG, as well as the \textbf{VMs embeddings}, enabling GATES to assess the relative importance of each VM to the ready task. Finally, these learned embeddings are concatenated to form the state embedding in GATES that comprehensively captures critical dynamics in CADWS, including the dynamic arrival of workflows, the changing DAG structures, and the dynamic variation in the number and status of VMs.

\subsubsection{The Decision-making Networks}
\begin{figure}[htbp]
\centering
\includegraphics[width=0.9\linewidth]{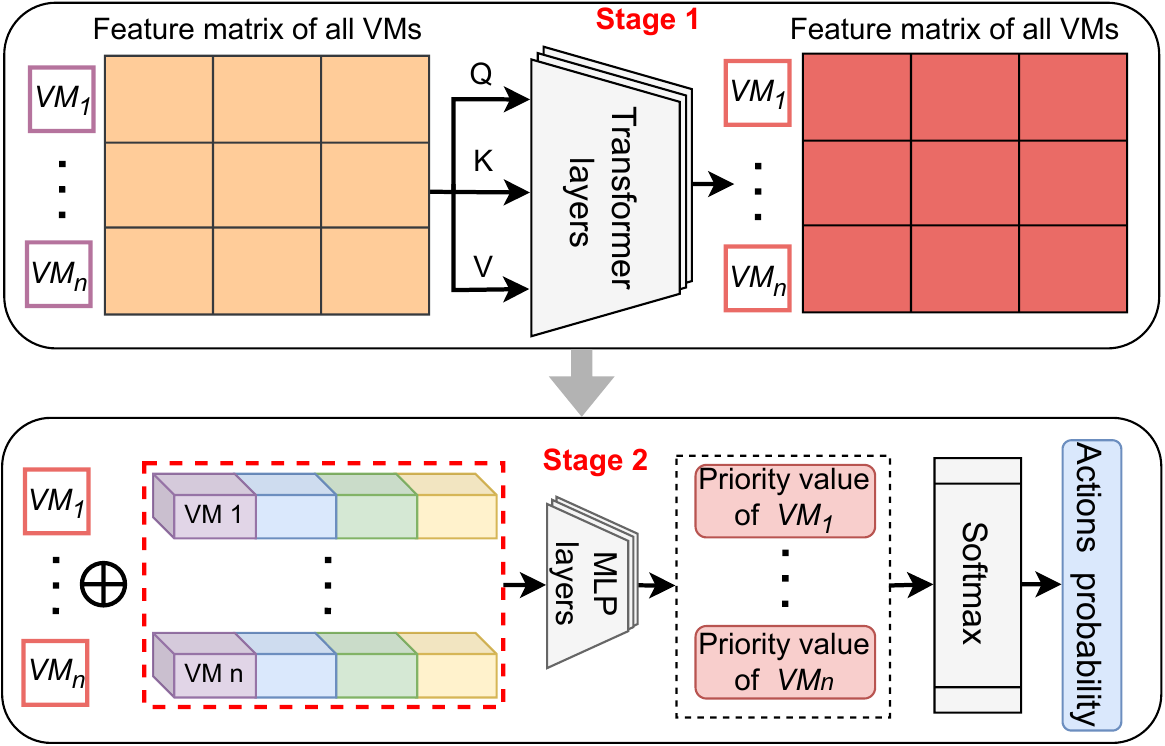}
\caption{The decision-making networks.}
\label{decision_networks}
\end{figure}

We design the decision-making networks, as shown in Figure~\ref{decision_networks}, to calculate the VM selection probabilities (i.e., the action distribution). Although the DHG of Figure~\ref{dynamic_graph} can capture the significance of dynamically changing VMs, an individual VM cannot reference the status information of others to determine its scheduling decisions. To address this, we employ a self-attention mechanism to learn global information among VMs, enabling each VM's feature embedding to account for the scheduling status of other VMs, thereby identifying the optimal VM to handle ready tasks. The computation process of the decision-making network consists of two stages: (1) the feature matrix of all VMs is first fed into a Transformer~\cite{vaswani2017attention}, where a self-attention mechanism captures the global information among VMs; (2) each VM's embedding is then concatenated with the state embedding to jointly determine the corresponding action score. The softmax function then computes the probability of selecting each VM based on these action scores. The detailed embedding learning process of Figure~\ref{HGAT} and Figure~\ref{decision_networks} is described further in Appendix B.

\subsection{Raw Features of CADWS} \label{subsec:raw_features}
The scheduling policy learned by GATES for CADWS relies on the state embedding derived from both the ready task (denoted as $O^{*}$) and all VMs at the current time step. The state embedding is constructed using three categories of raw features: \textbf{Workflow-related}, \textbf{Task-related}, and \textbf{VM-related} raw features. For any task ready for execution, these features are extracted from the current state of workflows, tasks, and VMs to construct the input to GATES, as summarized below.

\vspace{0.1cm}
\noindent
\textbf{Workflow-related raw features}:
\begin{itemize}
    \item \noindent \emph{Forthcoming successors}: the count of tasks dependent on $O^{*}$.
    \noindent \noindent \item \emph{Completion ratio}: the ratio of completed tasks ($C_{tasks}$) to total tasks ($T_{tasks}$) in $\mathcal{W}_{i}$, calculated as $C_{tasks}/T_{tasks}$.
    \item \noindent \emph{Estimated arrival rate}: an estimation of future workflow arrival rates based on the current execution status of $\mathcal{W}_{i}$.
\end{itemize}

\noindent
\textbf{Task-related raw features}:
\begin{itemize}
    \item \emph{Predecessor}: the number of predecessor tasks of each task in its corresponding DAG.
    \item \emph{Successor}: the number of successor tasks of each task in its corresponding DAG.
    \item \emph{Processing time}: the average duration required to execute a task on all types of VMs.
    \item \emph{Task size}: the workload of each task in DAG.
    \item \emph{Task deadline}: the task deadline is calculated using the method in \cite{wu2017deadline}. The deadline is assigned to each task, subject to the task size. A large task will be assigned a more relaxed task deadline.
    \item \emph{Task status}: the task status is represented by an integer: 0 for unscheduled, 1 for scheduled, and 2 for ready to be scheduled.
\end{itemize}

\noindent
\textbf{VM-related raw features}:
\begin{itemize}
    \item \emph{Task deadline feasibility}: an indicator that shows whether a VM can meet the deadline of $O^{*}$, assigned based on workflow size.
    \item \emph{Incurred cost}: the sum of the VM rental fee and SLA violation penalties for using the VM to execute $O^{*}$.
    \item \emph{Remaining rental time}: the remaining rental time on a VM after executing $O^{*}$.
    \item \emph{Fittest VM}: an indicator that shows whether a VM has the lowest incurred cost among all candidates while meeting the task deadline.
\end{itemize}

\subsection{Training Process via Evolution Strategy}
The performance of gradient-based DRLs is often influenced by hyperparameter choices and the design of reward functions, which can result in unstable or suboptimal learning outcomes~\cite{salimans2017evolution}. Evolution Strategy (ES) has been proposed as a robust alternative to mitigate these issues and achieve stable performance~\cite{khadka2018evolution,salimans2017evolution}. In this study, we employ ES to train the policy network denoted as $ \pi_{GATES} $. The pseudo-code of the training algorithm is presented in Algorithm~\ref{alg:algorithm}. The training process is described below:

1. \emph{Policy Sampling}: Let $ \hat{\theta} $ represent the current parameter of the policy $ \pi_{GATES} $. In each generation of ES, a population of $ N $ individuals is sampled. Specifically, for each individual $ \theta_i $, where $ i = 1, 2, \ldots, N $, $ \theta_i $ is drawn from an isotropic multivariate Gaussian distribution with mean $ \hat{\theta} $ and covariance $ \sigma^2 I $, i.e., $ \theta_i \sim \mathcal{N}(\hat{\theta}, \sigma^2 I) $. This can be expressed equivalently as $ \theta_i = \hat{\theta} + \sigma \epsilon_i $, where $ \epsilon_i \sim \mathcal{N}(0, I) $. Here, $ \theta_i $ denotes the parameters of the corresponding policy $ \pi_i $.

2. \emph{Fitness Evaluation}: The fitness of each sampled parameter $ \theta_i $ is evaluated as $ F(\theta_i) $, which corresponds to the total cost achieved by the policy $ \pi_i $ on a given CADWS problem instance. The fitness function is defined as $ F(\theta_i) = F(\hat{\theta} + \sigma \epsilon_i) = -\text{TotalCost}(\pi_i) $, where $ \text{TotalCost}(\pi_i) $ represents the overall cost of executing $ \pi_i $ as shown in Eq.~\eqref{Objective function}.

3. \emph{Policy Update}: To optimize the policy parameter $ \hat{\theta} $, ES maximizes the expected fitness $ \mathbb{E}_{\theta_i \sim \mathcal{N}(\hat{\theta}, \sigma^2 I)}[F(\theta_i)] $, effectively minimizing the total cost. The gradient is estimated as:
\begin{align}
\label{gradient_estimation}
\nabla_{\hat{\theta}} \mathbb{E}_{\theta_i \sim \mathcal{N}(\hat{\theta}, \sigma^2 I)}[F(\theta_i)] &= \nabla_{\hat{\theta}} \mathbb{E}_{\epsilon_i \sim \mathcal{N}(0, I)}[F(\hat{\theta} + \sigma \epsilon_i)]  \\
&\approx \frac{1}{N\sigma} \sum_{i=1}^{N} \{F(\hat{\theta}+\sigma\epsilon_{i})\epsilon_{i}\} \nonumber 
\end{align}

\begin{algorithm}[htpb!]
    \caption{The training process of Evolution Strategy}
    \label{alg:algorithm}
    \textbf{Input}: Population size: $N$, max number of generation: $Gen$, initial parameters of $\pi_{GATES}$: $\hat{\theta}$, initial learning rate: $\alpha$, and the Gaussian standard noise deviation: $\sigma$\\
    \textbf{Output}: The trained $\pi_{GATES}$
    \begin{algorithmic}[1] 
        \WHILE{the current number of generation $<=$ $Gen$}
        \STATE Randomly sample a CADWS training instance: $Pro$.
        \FOR{each individual ($i$=1,2,...) \textbf{in} $N$}
        \STATE Sample a $\epsilon_{i} \sim \mathcal{N}(0, I)$.
        \STATE The parameters of $\pi_{i}$ represented by individual $i$: $\theta_{i}=\hat{\theta}+\sigma\epsilon_{i}$
        \STATE Evaluate the fitness value of $F(\theta_{i})$ using Eq.~\eqref{Objective function} based on $Pro$
        \ENDFOR
        \STATE Estimate the policy gradient $\nabla_{\hat{\theta}} \mathbb{E}_{\theta_{i} \sim \mathcal{N}(\hat{\theta}, \sigma^{2}I)} F(\theta_{i})$ using Eq.~\eqref{gradient_estimation}.
        \STATE Update parameters of $\pi_{GATES}$: \\ $\hat{\theta} \leftarrow$ $\hat{\theta} + \alpha \frac{1}{N\sigma} \sum_{i=1}^{N} \{F(\hat{\theta}+\sigma\epsilon_{i})\epsilon_{i}\}$.
        \ENDWHILE
        \STATE \textbf{return} $\pi_{GATES}$
    \end{algorithmic}
\end{algorithm}
\section{Experiments}
We conduct experiments using a dedicated simulator designed for the CADWS problem~\cite{Yang2022icsoc,yanggraph}, leveraging Amazon EC2-based virtual machine (VM) configurations and four workflow patterns/DAGs: CyberShake, Montage, Inspiral, and SIPHT~\cite{deelman2015pegasus}, as shown in Appendix C. To reflect real-world resource heterogeneity, we employ six VM types with varying computational capacities and costs. Workflows are categorized into three scales, including small, medium, and large, based on the number of tasks, representing different levels of computational complexity. SLA deadlines are defined using a coefficient $\gamma \in \{1.00, 1.25, 1.50, 1.75, 2.00, 2.25\}$, guiding resource selection under varying time constraints. More details on training and testing can be found in Appendix D.

The proposed GATES method is compared against four baseline algorithms: \textbf{ProLis}~\cite{wu2017deadline}, \textbf{GRP-HEFT}~\cite{faragardi2019grp}, \textbf{ES-RL}~\cite{huang2022cost}, and \textbf{SPN-CWS}~\cite{shen2024cost}, spanning recently developed heuristic and DRL-based approaches. Key hyperparameters are configured to ensure fair comparison and computation efficiency. 

\subsection{Experiment Configuration}
\textbf{VM Configuration}: We have six VM types, ranging from m5.large to m5.12xlarge, with varying computational resources (e.g., vCPUs, memory) and hourly costs (e.g., \$0.096 to \$2.304). Each type of VM can be rented on demand to accommodate dynamically changing resource requirements. The detailed configuration of VMs is shown in Table~\ref{configuration_of_VMs}.
\begin{table}[htbp]
\centering
\begin{center}
\scalebox{0.7}{
\begin{tabular}{|c|c|c|c|c|c|}
\hline
\textbf{VM name} & \textbf{vCPU} & \textbf{Memory (GB)} & \textbf{Cost (\$ per hour)} \\ \hline
m5.large     & 2    & 8     & 0.096 \\ \hline
m5.xlarge    & 4    & 16    & 0.192 \\ \hline
m5.2xlarge   & 8    & 32    & 0.384 \\ \hline
m5.4xlarge   & 16   & 64    & 0.768 \\ \hline
m5.8xlarge   & 32   & 128   & 1.536 \\ \hline
m5.12xlarge  & 48   & 192   & 2.304 \\ \hline
\end{tabular}}
\end{center}
\vspace{-3mm}
\caption{The configuration of VMs.}
\label{configuration_of_VMs}
\end{table}

\textbf{Workflow Patterns}: As shown in Table~\ref{Workflow_Patterns}, workflows were categorized into three scales according to task size:
\begin{table}[htbp]
\centering
\begin{center}
\scalebox{0.7}{
\begin{tabular}{|c|c|c|c|c|}
\hline
\textbf{Workflow set} & \multicolumn{4}{c|}{\textbf{Name of pattern/number of task}} \\ \hline
\emph{Small}    & CyberShake/30   & Montage/25   & Inspiral/30   & SIPHT/30  \\ \hline
\emph{Medium}   & CyberShake/50   & Montage/50   & Inspiral/50   & SIPHT/60  \\ \hline
\emph{Large}    & CyberShake/100  & Montage/100  & Inspiral/100  & SIPHT/100  \\ \hline
\end{tabular}}
\end{center}
\vspace{-3mm}
\caption{The three workflow sets used in this study.}
\label{Workflow_Patterns}
\end{table}

\textbf{SLA Penalties}: SLA deadlines were defined using $\gamma$, which influenced resource selection. Larger SLA penalties encouraged renting cheaper VMs to manage cost while adhering to deadlines. The penalty weight $\beta$ was set to $0.24/hour$~\cite{shen2024cost}.

\textbf{Baseline Algorithms}: 
ProLis~\cite{wu2017deadline} and GRP-HEFT~\cite{faragardi2019grp} are heuristic methods. ES-RL~\cite{huang2022cost} is a DRL-based scheduling policy, developed based on OpenAI's code\footnote{https://github.com/openai/evolution-strategies-starter}. SPN-CWS~\cite{shen2024cost} is also a DRL-based method using the self-attention mechanism incorporated within the Transformer~\cite{vaswani2017attention}.
   
\textbf{Hyperparameters of GATES}: The hyperparameters of GATES mainly involve the parameters of HGAT (as shown in Figure~\ref{HGAT}), decision-making networks (as illustrated in Figure~\ref{decision_networks}), and the training method ES (see Algorithm~\ref{alg:algorithm}). 

In Figure~\ref{HGAT}, both stage 1 and stage 2 employ two GAT layers. The first layer has two attention heads and the second layer has one. The output dimensions $H$ of all GATs are 16. In Figure~\ref{decision_networks}, the Transformer layers consist of 2 encoder layers, each with 2 self-attention heads, an input and output dimension of 16, and a hidden layer dimension of 128. In Stage 2 of the decision-making network, the MLP consists of 4 layers with hidden dimensions of 128 and output dimension of 1, and uses ReLU as the activation function. For ES in Algorithm~\ref{alg:algorithm}, the hyperparameter settings are based on~\cite{huang2022cost} and~\cite{shen2024cost}. Specifically, we set the population size to 40, the maximum number of generations to 2000, the initial learning rate to 0.01, and the standard deviation of Gaussian noise to 0.05.

\subsection{Results and Analysis}
The results of all algorithms are presented in Table~\ref{Table:results}. The notation $ \langle 1.00, S \rangle $ indicates that $ \gamma = 1.00 $ and the algorithms are tested on the small-scale CADWS instance. ProLis and GRP-HEFT, being deterministic heuristic methods, do not have associated standard deviations. The symbols $ (+), (-), $ or $ (\approx) $ denote that the results are significantly better, worse, or statistically equivalent, respectively, compared to the corresponding algorithm. These comparisons are based on the Wilcoxon test with a significance level of 0.05.

As shown in Table~\ref{Table:results}, GATES achieves the lowest average total cost across all tested scenarios, encompassing various SLA coefficients ($\gamma$) as well as small (S), medium (M), and large (L) workflow sizes. It also demonstrates relatively small standard deviations, indicating more stable performance across multiple independent runs. Compared with existing approaches (ProLis, GRP-HEFT, ES-RL, and SPN-CWS), GATES exhibits notable cost advantages in multiple scenarios. For instance, under the $\langle1.50, M\rangle$ scenario, the total cost is reduced to 196.90, which is notably lower than the 276.09 achieved by SPN-CWS and also surpasses the results of other algorithms. Even under tight $\gamma$ (=1) and with large workflow size, GATES consistently maintains the lowest cost, highlighting its adaptability across both relaxed and stringent SLA constraints. A Wilcoxon rank-sum test ($p<$ 0.05) based on 30 independent runs confirms the statistical significance of these differences, indicating that GATES significantly outperforms the current state-of-the-art methods for workflows with varying patterns and scales. Considering the average cost, variance, and statistical evidence, GATES offers a distinct combined advantage in reducing overall cost while preserving performance stability, providing a more practical approach for solving CADWS.
\begin{table*}[htbp]
\label{Table:results}
\centering
\begin{center}
\scalebox{0.75}{
\begin{tabular}{|c|c|c|c|c|c|}
\hline
Scenarios & ProLis & GRP-HEFT & ES-RL & SPN-CWS & GATES (our) \\ \hline
\(\langle 1.00, S \rangle\) & 773.69   & 1685.01(-)  & 223.68(51.51)(+)(+)    & 158.28(14.45)(+)(+)(+)       & \textbf{143.95(18.94)}(+)(+)(+)(+) \\  
\(\langle 1.00, M \rangle\) & 1829.25  & 2867.64(-)  & 422.26(100.92)(+)(+)   & 308.93(75.88)(+)(+)(+)       & \textbf{225.41(37.21)}(+)(+)(+)(+) \\  
\(\langle 1.00, L \rangle\) & 3641.79  & 5873.20(-)  & 829.64(262.91)(+)(+)   & 524.83(163.15)(+)(+)(+)      & \textbf{324.93(65.65)}(+)(+)(+)(+) \\ \hline
\(\langle 1.25, S \rangle\) & 923.15   & 1967.31(-)  & 320.75(83.69)(+)(+)    & 143.12(15.73)(+)(+)(+)       & \textbf{124.14(7.10)}(+)(+)(+)(+) \\  
\(\langle 1.25, M \rangle\) & 2068.78  & 3578.04(-)  & 565.53(213.00)(+)(+)   & 272.77(60.65)(+)(+)(+)       & \textbf{193.26(24.06)}(+)(+)(+)(+) \\ 
\(\langle 1.25, L \rangle\) & 4213.29  & 6868.22(-)  & 989.79(458.04)(+)(+)   & 496.63(155.02)(+)(+)(+)      & \textbf{300.19(76.29)}(+)(+)(+)(+) \\ \hline
\(\langle 1.50, S \rangle\) & 901.40   & 1963.39(-)  & 270.93(67.20)(+)(+)    & 144.13(18.97)(+)(+)(+)       & \textbf{121.56(7.11)}(+)(+)(+)(+) \\ 
\(\langle 1.50, M \rangle\) & 2035.52  & 3567.97(-)  & 505.44(222.45)(+)(+)   & 276.09(60.63)(+)(+)(+)       & \textbf{196.90(33.40)}(+)(+)(+)(+) \\ 
\(\langle 1.50, L \rangle\) & 4066.84  & 6863.62(-)  & 886.87(380.26)(+)(+)   & 511.69(161.85)(+)(+)(+)      & \textbf{316.23(99.12)}(+)(+)(+)(+) \\ \hline
\(\langle 1.75, S \rangle\) & 804.33   & 1959.71(-)  & 241.09(51.43)(+)(+)    & 137.36(14.10)(+)(+)(+)       & \textbf{118.45(4.85)}(+)(+)(+)(+)  \\
\(\langle 1.75, M \rangle\) & 1977.10  & 3560.22(-)  & 491.97(229.48)(+)(+)   & 263.12(57.61)(+)(+)(+)       & \textbf{195.25(35.83)}(+)(+)(+)(+) \\ 
\(\langle 1.75, L \rangle\) & 3898.42  & 6854.40(-)  & 828.06(359.74)(+)(+)   & 502.67(155.64)(+)(+)(+)      & \textbf{315.84(92.48)}(+)(+)(+)(+) \\ \hline
\(\langle 2.00, S \rangle\) & 789.71   & 1957.25(-)  & 225.70(50.89)(+)(+)    & 129.75(12.72)(+)(+)(+)       & \textbf{113.48(5.30)}(+)(+)(+)(+)   \\
\(\langle 2.00, M \rangle\) & 1921.42  & 3554.07(-)  & 427.16(213.71)(+)(+)   & 251.83(52.01)(+)(+)(+)       & \textbf{185.55(30.66)}(+)(+)(+)(+) \\  
\(\langle 2.00, L \rangle\) & 4024.03  & 6847.49(-)  & 780.77(408.14)(+)(+)   & 497.65(158.52)(+)(+)(+)      & \textbf{322.01(110.05)}(+)(+)(+)(+) \\ \hline
\(\langle 2.25, S \rangle\) & 710.18   & 1954.71(-)  & 201.54(50.56)(+)(+)    & 126.18(13.02)(+)(+)(+)       & \textbf{110.90(5.70)}(+)(+)(+)(+)   \\
\(\langle 2.25, M \rangle\) & 1924.27  & 3551.31(-)  & 374.96(190.21)(+)(+)   & 238.35(48.26)(+)(+)(+)       & \textbf{179.17(24.54)}(+)(+)(+)(+) \\ 
\(\langle 2.25, L \rangle\) & 3957.08  & 6842.88(-)  & 678.94(396.79)(+)(+)   & 475.46(139.99)(+)(+)(+)      & \textbf{318.19(112.21)}(+)(+)(+)(+) \\ \hline
\end{tabular}}
\end{center}
\vspace{-3mm}
\caption{Average (standard deviation) total cost of each algorithm over 30 independent runs.}\label{Table:results}
\end{table*}

\subsection{Trade-off between VM Fees and SLA Penalties}
\begin{figure}[htbp]
\centering
\includegraphics[width=\linewidth]{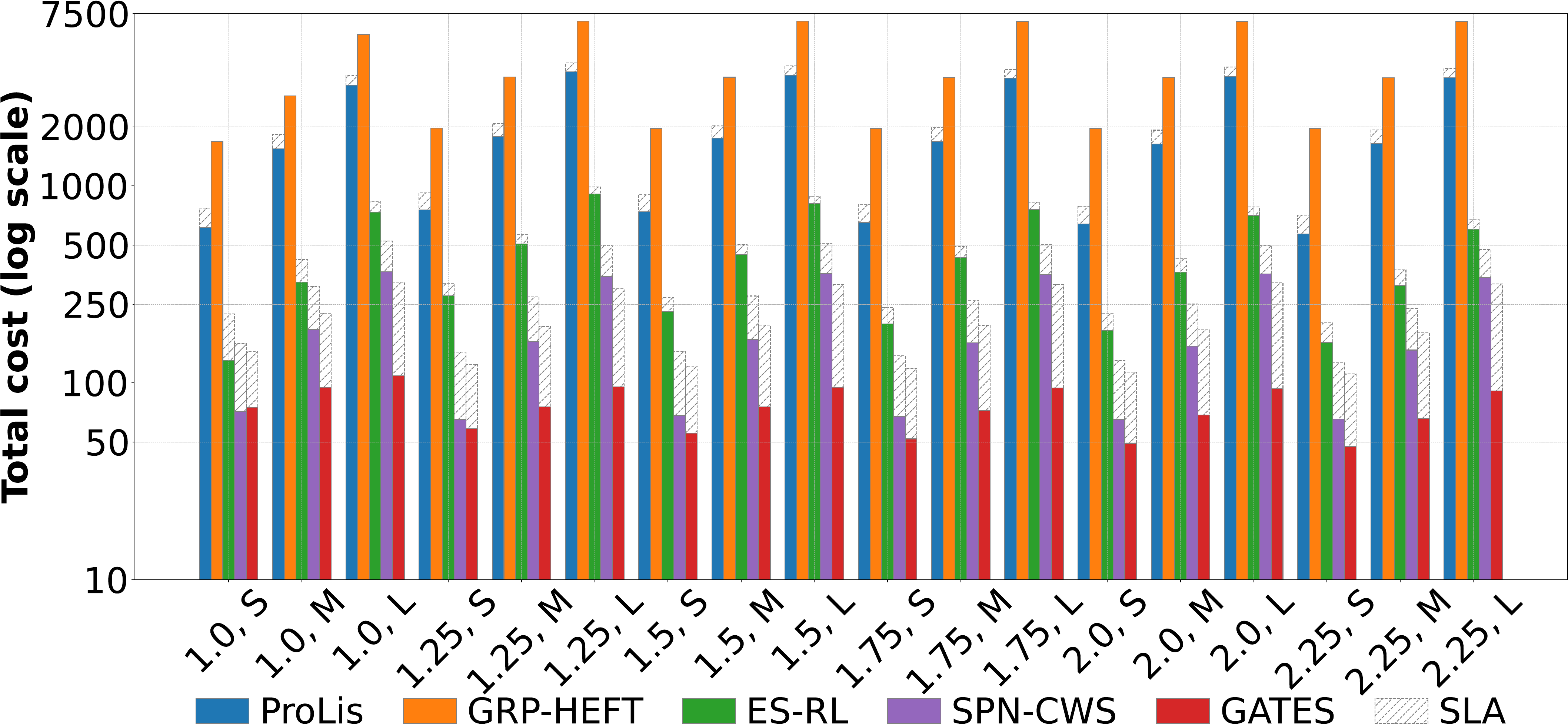} 
\caption{The trade-off between  VM rental fees and SLA penalty.}
\label{results_bar}
\end{figure}

Figure~\ref{results_bar} compares the VM rental fees and corresponding SLA penalty costs for all algorithms under various scenarios. The y-axis is plotted on a logarithmic scale for better visualization, with each algorithm’s bar representing VM fees and the diagonally striped portion indicating SLA penalties. GRP-HEFT enforces strict SLA deadlines, leading to high VM fees but negligible penalties, while ProLis exhibits a similar pattern. ES-RL employs reinforcement learning for adaptive allocation, obtaining moderately high fees and relatively low penalties. SPN-CWS focuses on reducing VM rental costs by using cheaper VMs at the price of high SLA penalties. In comparison, GATES demonstrates strong capabilities in balancing VM rental fees and SLA penalties to effectively reduce the total cost. Although GATES incurs slightly higher SLA penalties than ES-RL and SPN-CWS in some scenarios, it significantly lowers the total cost by effectively leveraging cheap VMs for task scheduling.  

Overall, Figure~\ref{results_bar} illustrates distinct trade-offs between cost and SLA compliance. While GRP-HEFT and ProLis favor meeting tight deadlines at the expense of higher VM fees, and SPN-CWS accepts greater penalties to minimize costs, GATES achieves a strong balance between these objectives by keeping both VM fees and SLA penalties within a favorable range to reduce the total cost.

\subsection{Convergence}  
\begin{figure}[htbp]
\centering
\small
\subfigure{\includegraphics[width=.48\columnwidth]{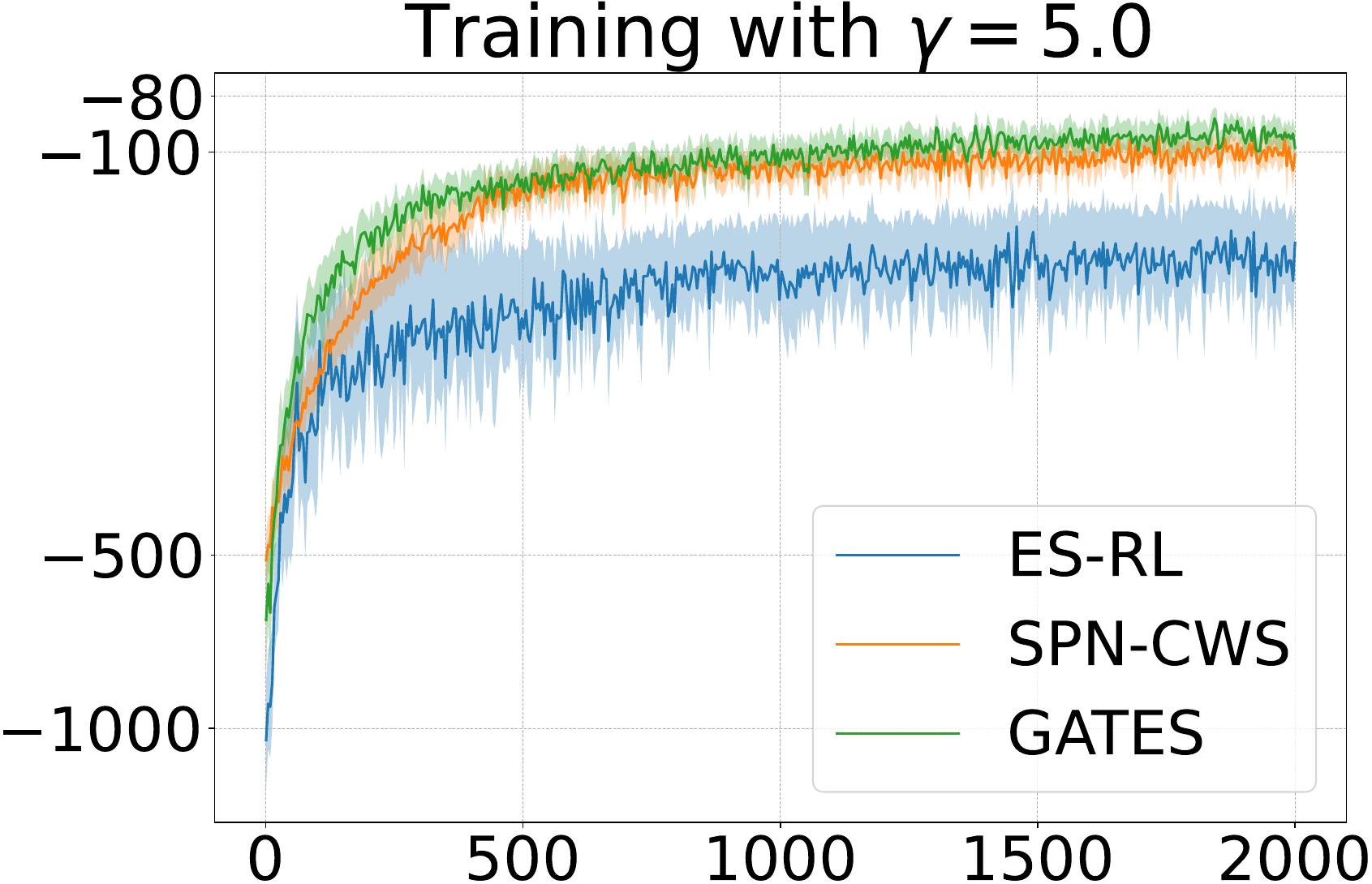}}
\subfigure{\includegraphics[width=.48\columnwidth]{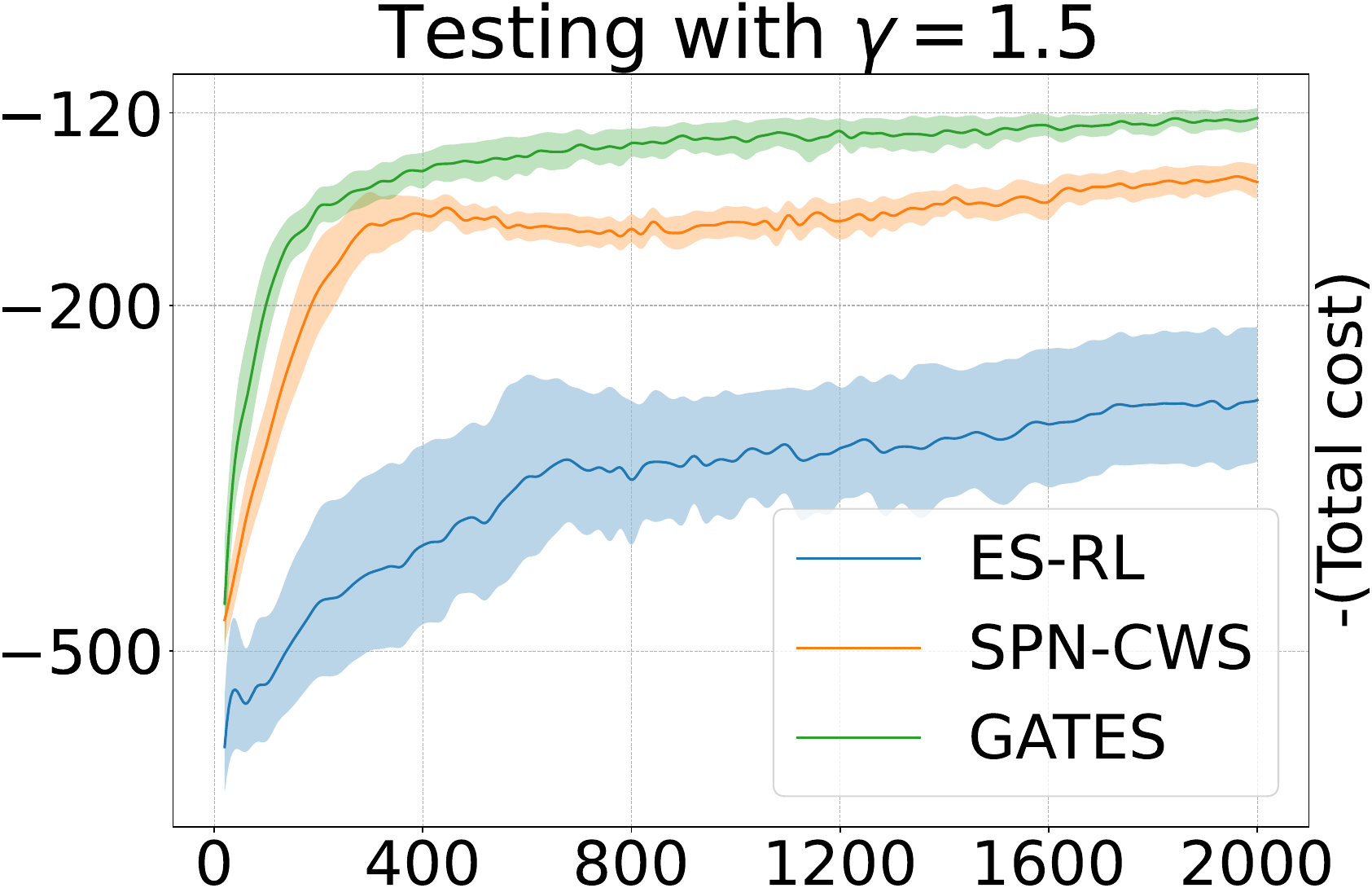}}
\caption{The convergence on the small-scenario instance. The horizontal axis denotes the training generation, while the vertical axis indicates the negative total cost.}
\label{fig:difference in train and test}
\end{figure}

As shown in Figure~\ref{fig:difference in train and test}, all three DRL-based algorithms (ES-RL, SPN-CWS, and GATES) converge during both training and testing, but their final performances vary significantly. GATES achieves superior results compared to the other algorithms, with its total cost curve significantly lower, indicating its ability to learn better scheduling policies in complex dynamic environments to reduce the total scheduling cost of CADWS. SPN-CWS ranks second, with convergence levels lower than GATES but better than ES-RL. In contrast, ES-RL exhibits lower curves with higher fluctuations, suggesting poorer policy quality and less stability.

From the perspective of confidence intervals, both GATES and SPN-CWS demonstrate smaller fluctuations in training and testing curves, showing greater robustness to random seeds and environmental disturbances. Although some oscillations occur in the early iterations, their overall convergence variance is notably better than that of ES-RL. In comparison, ES-RL falls short in both solution quality and stability. Overall, GATES excels in solution quality, with its learned scheduling policies offering superior stability and significantly outperforming the other two DRL algorithms. This analysis further validates that the DHG constructed in GATES effectively captures the dynamics of CADWS, and the designed HGAT network extracts meaningful state embeddings to learn efficient scheduling policies.

\section{Conclusion}
This paper addresses the complex characteristics of CADWS in cloud environments, such as dynamic workflow arrivals and resource variations, by proposing a novel DRL method called GATES, which integrates GAT and ES. GATES learns task-level topological dependencies and VM importance to dynamically allocate tasks to optimal VMs while leveraging ES to avoid the challenges of complex reward design and hyperparameter tuning. Experimental results demonstrate that GATES can learn an efficient and stable scheduling policy, significantly outperforming several existing methods.

\section*{Ethical Statement}
There are no ethical issues. The name of ``GATES" is used coincidentally and has no relation to~\cite{ning2020generic} in terms of problem setting or technical content.

\section*{Acknowledgments}
The authors would like to express their gratitude to the anonymous reviewers for their valuable feedback and constructive suggestions. The authors also acknowledge Victoria Huang, Chen Wang, Yifan Yang, and their collaborators for their prior works, on which the simulation environment used in this study is based. This research is partially supported by Grant VUW-FSRG-
410114, administered by Victoria University of Wellington.


\bibliographystyle{named}
\bibliography{ijcai25}

\clearpage
\appendix

\begin{center}
	\Large
	Appendix
\end{center}

\section{The details of workflow patterns}\label{app:DAG}
\begin{figure}[htbp]
\centering
\includegraphics[width=\linewidth]{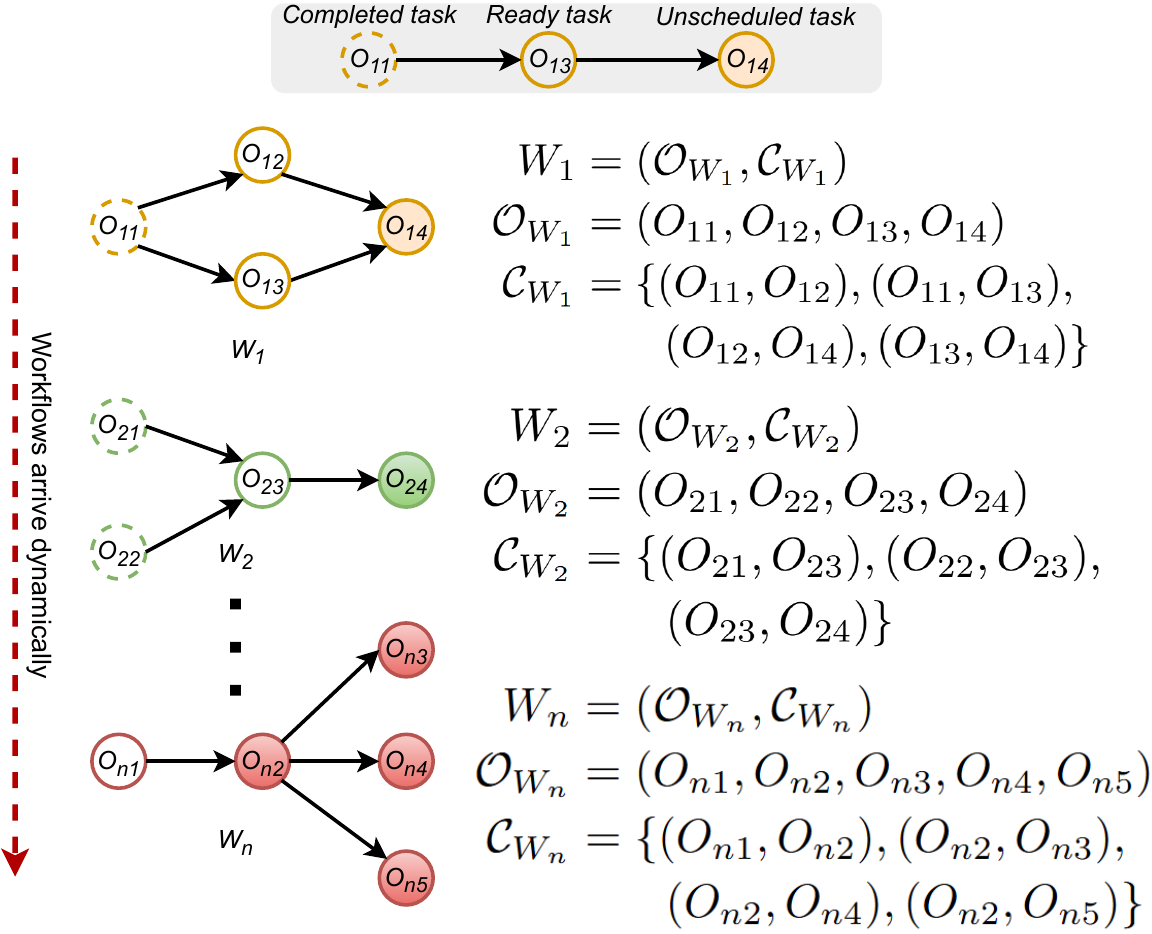} 
\caption{The sketch of workflow patterns/DAGs}
\label{DAG}
\end{figure}

A workflow $ W_i $ is modeled as a Directed Acyclic Graph (DAG), denoted as $ W_i = (\mathcal{O}_{W_i}, \mathcal{C}_{W_i}) $, as illustrated in Figure~\ref{DAG}. Here, $ \mathcal{O}_{W_i} = \{ O_{i1}, O_{i2}, \ldots, O_{in_i} \} $ represents the set of nodes, where each node $ O_{ij} $ corresponds to a specific task in $ W_i $ that requires execution. The edge set $ \mathcal{C}_{W_i} $ defines the directed dependencies between tasks, such that an edge $ (O_{ij}, O_{ik}) \in \mathcal{C}_{W_i} $ imposes a precedence constraint, meaning task $ O_{ij} $ must complete before $ O_{ik} $ can begin.

The execution of tasks within the workflow adheres to the topological order defined by the DAG. A task becomes eligible for scheduling or execution only when all its predecessor tasks have been completed; such a task is termed as a \emph{ready task}. \textcolor{blue}{In this study, we do not assume any workflow or task priority. Following the \emph{no-delay principle}~\cite{zhu2016fault}, every task is scheduled immediately when it becomes ready. The broker always allocates resources first to the earliest known ready task.}

\section{The embedding process of GATES}\label{appendix:embedding_GATES}
The embedding learning process in GATES consists of two components: HGAT and the decision-making networks. We will introduce the embedding learning process of these two components separately.

\textbf{HGAT}: The embedding process of HGAT consists of two main stages. In Stage 1, \textcolor{blue}{a Graph Attention Network (GAT) is used to learn feature embeddings for each task from the workflow DAG. The \emph{Ready-task embedding 1} in Fig.~\ref{HGAT} refers to the embedding of the current ready task. In addition, the learned feature embeddings of all tasks are aggregated via mean pooling to obtain the \emph{All taks embedding} that captures the embeddings of the whole workflow DAG.}

\textcolor{blue}{Concretely, the GAT attention coefficient between a task node $O_{ni}$ and one of its neighbors $O_{nj}$ is computed as:
\begin{equation}
\alpha_{ij} = \frac{\exp\!\Bigl(\mathrm{LR}\!\bigl(\mathbf{a}^\top 
[\mathbf{W}\,\mathbf{h}_{O_{ni}}\;\|\;\mathbf{W}\,\mathbf{h}_{O_{nj}}]
\bigr)\Bigr)}{\sum_{O_{nk} \in \mathcal{N}(O_{ni})}\exp\!\Bigl(\mathrm{LR}\!\bigl(\mathbf{a}^\top[\mathbf{W}\,\mathbf{h}_{O_{ni}}\;\|\;\mathbf{W}\,\mathbf{h}_{O_{nk}}]\bigr)\Bigr)},
\label{eq:gat-attention}
\end{equation}}
\noindent
\textcolor{blue}{The updated embedding of node $O_{ni}$ is then obtained via a weighted aggregation of its neighbors:
\begin{equation}
\mathbf{h}_{O_{ni}}' = \sigma\!\Bigl(\sum_{O_{nj} \in \mathcal{N}(O_{ni})} \alpha_{ij} \, \mathbf{W} \, \mathbf{h}_{O_{nj}}\Bigr),
\label{eq:gat-update}
\end{equation}}
\noindent
\textcolor{blue}{The \emph{All tasks embedding} is computed by averaging all task node embeddings from the DAG:
\begin{equation}
\mathbf{h}_{W_n} = \mathrm{mean}\Bigl(\sum_{O_{ni} \in \mathcal{O}_{W_n}} \mathbf{h}_{O_{ni}}'\Bigr),
\label{eq:workflow-embedding}
\end{equation}}

\textcolor{blue}{In Eq.~\eqref{eq:gat-attention}, $\alpha_{ij}$ is the GAT attention coefficient from $O_{ni}$ to $O_{nj}$, $O_{ni}$ denotes the $i$-th task node of workflow $\mathcal{W}_n$, and $\mathcal{N}(O_{ni})$ represents all its neighboring task nodes. The vector $\mathbf{h}_{O_{ni}} \in \mathbb{R}^{d}$ denotes the input feature of the task node $O_{ni}$, and $\mathbf{h}_{O_{nj}} \in \mathbb{R}^{d}$ represents the input feature of its neighboring task node $O_{nj}$. The specific features used for each task node are listed in Section~\ref{subsec:raw_features}. $\mathbf{W}$ is a learnable linear transformation matrix. The vector $\mathbf{a}$ is a trainable attention vector, and $\|$ indicates feature concatenation. The function $\mathrm{LR}(\cdot)$ is the LeakyReLU activation. In Eq.~\eqref{eq:gat-update}, the $\sigma(\cdot)$ denotes the ReLU nonlinear activation function. In HGAT, the hidden or output dimensions are all set to $H$. The resulting vector $\mathbf{h}_{O_{ni}}'\in \mathbb{R}^{H}$ is the updated node embedding of $O_{ni}$. In Eq.~\eqref{eq:workflow-embedding}, $\mathcal{O}_{W_n}$ is the set of all task nodes in workflow $\mathcal{W}_n$, and $mean(\cdot)$ is the mean pooling function to average the embeddings of all nodes. The vector $\mathbf{h}_{W_n}\in \mathbb{R}^{H}$ is the average of all node embeddings in this workflow. When $O_{ni}$ corresponds to the currently ready task $O_{ni}^{*}$ awaiting scheduling, its embedding $\mathbf{h}_{O_{ni}}'$ is referred to as the \emph{Ready-task embedding 1}, and the vector $\mathbf{h}_{W_n}$ is referred to as the \emph{All nodes embedding}.}

\textcolor{blue}{In stage 2, another multi-layer GAT is applied to extract embeddings from the VM-graph part of the DHG. The embedding of the ready task is referred to as the \emph{Ready-task embedding 2}, and the embeddings of all VMs are collectively denoted as the \emph{VMs embedding}. In the VM-graph, the features of the ready task cover the workflow-related features, listed in Section~\ref{subsec:raw_features}. The raw features of VMs are the VM-related features. The attention coefficients between connected nodes in the VM-graph are computed using the same formulation as in Eq.~\eqref{eq:gat-attention}, and the embedding of each node is updated through weighted aggregation of its neighbors following Eq.~\eqref{eq:gat-update}. After this process, the embedding of the ready task is denoted as \emph{Ready-task embedding 2} ($\mathbf{h}_{O_{ni}}''\in \mathbb{R}^{H}$), and the embeddings of all VMs are combined to form the \emph{VMs embedding} ($\mathbf{X}_{vm}'\in \mathbb{R}^{n \times H}$). The final learned state embedding $\mathbf{S}_{O_{ni}}\in \mathbb{R}^{n \times 4H}$ is formed via dimension-wise concatenation (denoted by $\oplus$), integrating $\mathbf{h}_{O_{ni}}'$, $\mathbf{h}_{W_{n}}$, $\mathbf{h}_{O_{ni}}''$, and $\mathbf{X}_{vm}'$ into a comprehensive state embedding for the current ready task $O_{ni}$ as shown in Eq.~\eqref{eq:state_embedding}.
\begin{equation}
\label{eq:state_embedding}
\mathbf{S}_{O_{ni}} = \mathbf{h}_{O_{ni}}' \oplus \mathbf{h}_{W_{n}} \oplus \mathbf{h}_{O_{ni}}'' \oplus \mathbf{X}_{vm}',
\end{equation}}
This unified state embedding comprehensively captures task dependencies and dynamic VM characteristics in CADWS.

\textbf{Decision-making networks}: In the first stage, we feed the feature matrix $\textcolor{blue}{X_{vm}} \in \mathbb{R}^{n \times m}$, whose rows $n$ represent the raw features of the $n$-th VM, into a multi‐head self‐attention (MHSA) module from Transformer, as shown in Eq.~\ref{eq:detail_of_MHSA}. This stage computes \emph{query} ($Q$)/\emph{key} ($K$)/\emph{value} ($V$) matrices and learns global context information across all VMs. 
\begin{align} 
\label{eq:detail_of_MHSA}
Q = XW_Q, K &= XW_K, V = XW_V, \\ \nonumber
\mathrm{Attn}(Q,K,V) &= \mathrm{softmax}\!\Bigl(\tfrac{QK^\top}{\sqrt{d_k}}\Bigr)\,V, \\ \nonumber
\mathrm{MHSA}(X) &= \bigl[\mathrm{head}_1 \,\|\,\dots\,\|\,\mathrm{head}_h\bigr]W^O,
\;\; \\ \nonumber
\mathrm{head}_i &= \mathrm{Attn}\!\bigl(XW_Q^i,\;XW_K^i,\;XW_V^i\bigr), \nonumber
\end{align}

In the second stage, each refined VM embedding is concatenated with the learned state embedding from HGAT, and then passed through an MLP to produce action scores, which are normalized via softmax to yield action probabilities. The detailed process is shown in the following:
\begin{enumerate}
\item $\textbf{MHSA Update:}\quad
X' = \mathrm{MHSA}(X),$
\item $\textbf{Concatenated Embedding:}\quad
\hat{\mathbf{x}}_i = [\mathbf{x}'_i;\mathbf{S}_{O_{ni}}],$
\item $\textbf{Action Scores:}\quad
\alpha_i = \mathrm{MLP}(\hat{\mathbf{x}}_i),$
\item $\textbf{Softmax Probability:}\quad
p_i = \frac{\exp(\alpha_i)}{\sum_{j=1}^k \exp(\alpha_j)},$
\item $\textbf{Action Selection:} \\ \quad
\mathrm{action} = \begin{cases} \arg\max_i \alpha_i,& \text{(greedy)}\\
\text{sample from }p_i,& \text{(sampling)}
\end{cases}$
\end{enumerate}

Here, $W_Q, W_K, W_V, W_Q^i, W_K^i, W_V^i, \text{ and } W^O$ of Eq.~\ref{eq:detail_of_MHSA} are trainable weight matrices, and $\mathbf{x}_i'\in\mathbb{R}^{d}$ denotes the refined embedding of VM $i$ obtained from $X'$. $X$ is the original feature matrix of all VMs. $\mathbf{S}_{O_{ni}}$ is the State embedding from HGAT, as shown in Eq.~\ref{eq:state_embedding}. Additionally, a learned workflow feature through an MLP will also be concatenated with them. The symbol $\|$ denotes concatenation, while $[\cdot;\,\cdot]$ is the operation of feature‐wise concatenation. $\alpha_i$ is the action score, and $p_i$ is the normalized action probability. In this study, we select the VM action greedily.

\section{Workflow patterns experimented}\label{app:DAG_types}
\begin{figure}[htbp]
\centering
\includegraphics[width=\linewidth]{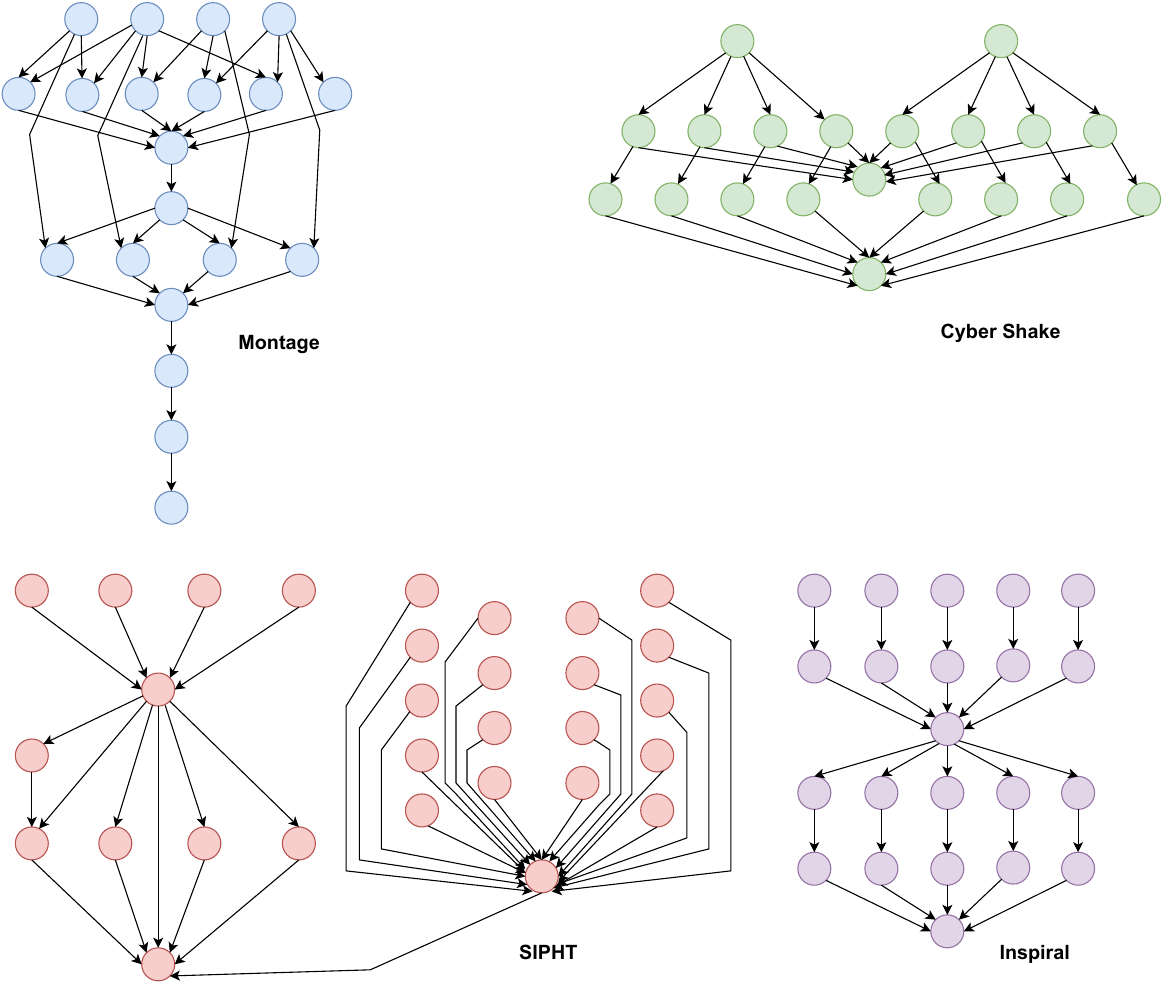} 
\caption{The workflows studied in this paper}
\label{fg:patterns}
\end{figure}

In this study, we adopt a commonly used experimental framework in the cloud computing field~\cite{shen2024cost,huang2022cost}. This framework leverages four extensively studied workflow patterns~\cite{deelman2015pegasus}, which are illustrated in Figure~\ref{fg:patterns}. Each workflow pattern provides detailed specifications, including task workloads $ WL_{ij} $ (as detailed in Section~\ref{sec:problem_definition}) and the dependency structures between tasks. For additional reference, readers can consult the Pegasus user guide~\footnote{https://pegasus.isi.edu/documentation/user-guide/introduction.html} available at~\cite{deelman2015pegasus}. In our experiments, these workflow patterns arrive dynamically, following a Poisson distribution characterized by parameter $ \lambda $ as referred to Eq.~\ref{Objective function}. This setup aligns with the standard configurations widely adopted in prior research~\cite{shen2024cost,huang2022cost}.

\textcolor{blue}{
\section{The details of training and testing}\label{app:exper_settings}
The strictness of the Service Level Agreement (SLA) deadline is governed by a parameter $\gamma$. For training, all methodologies undergo a learning phase under a less restrictive condition where $\gamma=5.0$. Conversely, the evaluation phase employs more demanding conditions, with $\gamma$ taking a series of values in $\{1.00, 1.25, 1.50, 1.75, 2.00, 2.25\}$, to rigorously assess the generalization capabilities of the acquired policies.}

\textcolor{blue}{Each experimental instance consists of 30 workflows, randomly drawn, with their arrival times following a Poisson distribution ($\lambda = 0.01$) to simulate realistic dynamic workflow arrivals. The scale of these instances is classified into \textit{small}, \textit{medium}, and \textit{large}, contingent on whether all constituent workflows are chosen from their respective subsets as detailed in Table~\ref{Workflow_Patterns}. The training process is only conducted on small-scale instances, whereas the evaluation process involves 30 randomly selected instances spanning all three scales. To guarantee the statistical significance of the findings, all presented results represent averages obtained from 30 independent tests.
}

\end{document}